\documentclass{article}
\usepackage{spconf,amsmath,epsfig, etoolbox}

\providecommand{\doi}[1]{doi: {\footnotesize \href{http://dx.doi.org/#1}{\path{#1}}}}

\usepackage[pdftex=true,breaklinks=true,hidelinks=true,colorlinks=true,citecolor=blue]{hyperref}

\usepackage[square,numbers]{natbib}

\let\OLDthebibliography\thebibliography
\renewcommand\thebibliography[1]{
  \OLDthebibliography{#1}
  \setlength{\parskip}{0pt}
  \setlength{\itemsep}{0pt plus 0.3ex}
}

\title{Leveraging Multi-Temporal Sentinel 1 and 2 Satellite Data for Leaf Area Index Estimation with Deep Learning}
\name{Clement Wang*, Antoine Debouchage*, Valentin Goldité*, Aurélien Wery, Jules Salzinger**}

\address{*CentraleSupélec - University of Paris-Saclay, **Austrian Institute of Technology - \textit{Vienna, Austria}}

\begin{document}
%
\maketitle
\begin{abstract}

The Leaf Area Index (LAI) is a critical parameter to understand ecosystem health and vegetation dynamics. In this paper, we propose a novel method for pixel-wise LAI prediction by leveraging the complementary information from Sentinel 1 radar data and Sentinel 2 multi-spectral data at multiple timestamps. Our approach uses a deep neural network based on multiple U-nets tailored specifically to this task. To handle the complexity of the different input modalities, it is comprised of several modules that are pre-trained separately to represent all input data in a common latent space.  Then, we fine-tune them end-to-end with a common decoder that also takes into account seasonality, which we find to play an important role. Our method achieved 0.06 RMSE and 0.93 R² score on publicly available data. We make our contributions available\footnote{\href{https://github.com/valentingol/LeafNothingBehind}{https://github.com/valentingol/LeafNothingBehind}} for future works to further improve on our current progress.

\end{abstract}

\section{INTRODUCTION}
\label{sec:intro}

The Leaf Area Index (LAI) is a fundamental vegetation parameter that quantifies the total area of leaves per unit ground area. It serves as a key indicator of plant productivity, energy exchange processes, and overall health of the ecosystem \cite{b1}. Accurate estimation of the LAI is essential for various applications, including ecological modeling, crop yield prediction, carbon cycle assessment, and climate change studies.

Traditionally, the LAI has been estimated using labor-intensive and time-consuming methods, such as destructive sampling or indirect measurements based on allometric equations \cite{b2} \cite{b3}. While these approaches provide valuable insights, they are limited in their spatial coverage and accuracy. 

As a result, remote sensing data from satellite platforms have gained significant attention as a valuable source for LAI predictions. Satellite-based observations offer the advantage of providing repetitive coverage over large areas, enabling the assessment of LAI dynamics at regional to global scales. 
However, despite their many advantages, satellite data face significant challenges, notably the interference caused by clouds in accurate LAI estimation. Cloud cover poses a substantial obstacle to LAI assessment, as it obstructs direct measurements and reduces the quality and availability of cloud-free observations. Overcoming this challenge is critical to leveraging the full potential of satellite data for comprehensive and reliable LAI prediction. 

Our approach focuses on pixel-wise LAI prediction using deep learning. To the best of our knowledge, this study represents a pioneering investigation into the prediction of LAI by harnessing the combined information from Sentinel 1 and Sentinel 2 data at multiple timestamps with a deep neural network.

\begin{figure*}
  \centering
  \includegraphics[scale=0.4]{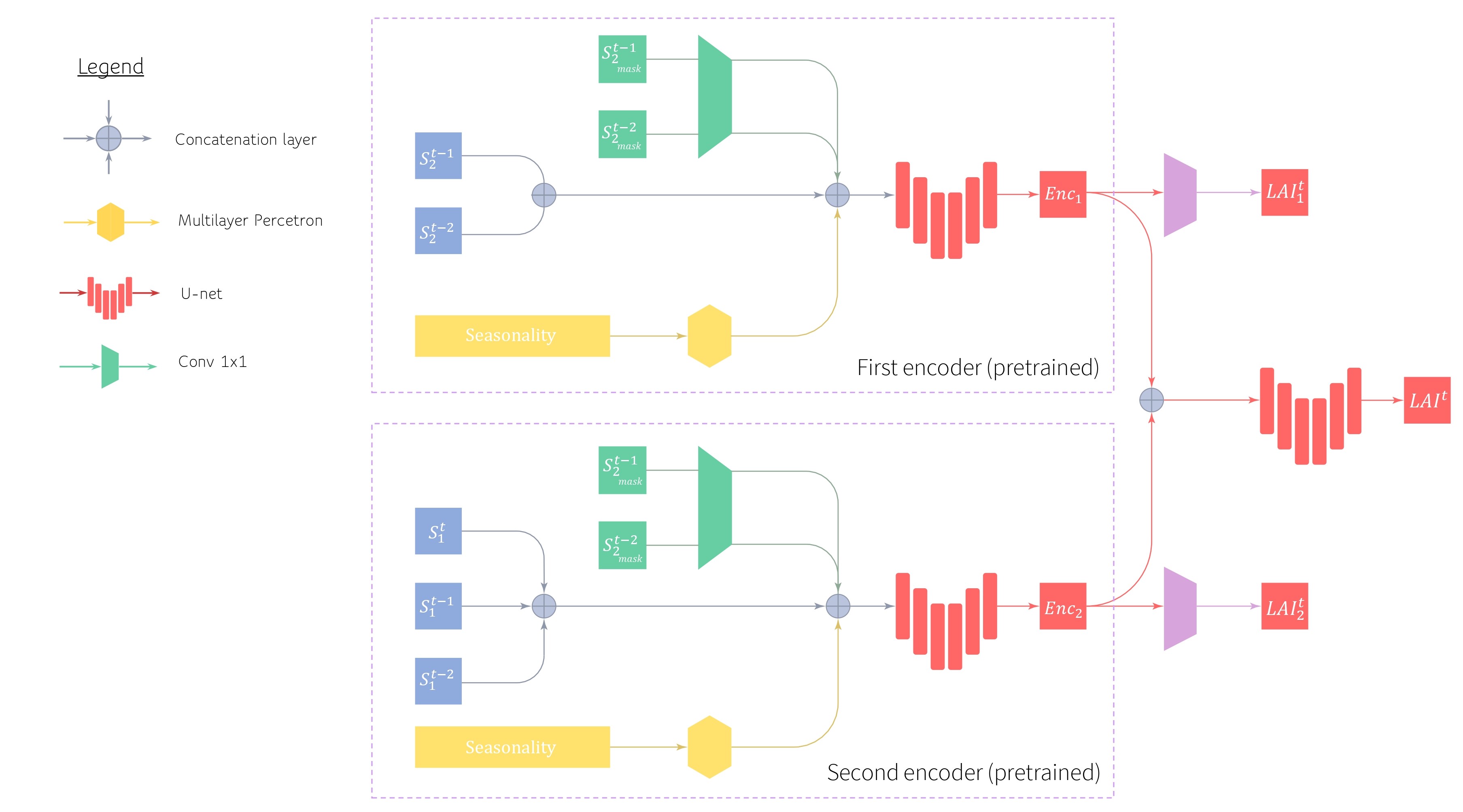}
  \caption{\label{strontium} Proposed architecture with two parallel encoders and one decoder}
  \label{archi_figure}
\end{figure*}


\section{RELATED WORKS}
\label{sec:format}

Satellite data has gained popularity due to its convenience in LAI prediction, offering significant advantages over direct methods or allometric approaches \cite{b2} \cite{b3}. 
Moreover, as demonstrated by \cite{b4} and \cite{b5}, satellite data exhibit comparable performance to traditional alternatives. In certain cases, it can even outperform them in explainability, particularly when supplemented with additional metadata such as terrain variables.

Most studies on LAI prediction from satellite data have primarily focused on specific settings, often limited to a single type of crop and to the optical data from Sentinel 2 or Lansat  8. These studies commonly employ various methods, including multi-regression \cite{b5} \cite{b6} \cite{b7} \cite{b8}, Fully connected neural networks \cite{b8} \cite{b9}, Bayesian networks \cite{b10}, as well as classical machine learning techniques \cite{b11} \cite{b12} such as gradient boosting, Gaussian process, support vector machines (SVM), and random forest. \cite{b13} adopts Gaussian processes on Sentinel 2 and Sentinel 1 data at multiple timestamps, resembling our approach. A frequent limitation in prior work is that each pixel is treated independently, overlooking the valuable contextual information from neighboring pixels. This disregard for the global context surrounding each pixel is a missed opportunity that warrants further exploration and consideration.

In parallel, cloud removal has garnered significant attention within the realm of deep learning research. Numerous studies have explored the development of robust methods for cloud removal on satellite data using deep learning techniques. In particular, more powerful convolutional architectures have been investigated, capitalizing on the convolutional inductive bias that takes advantage of neighboring pixel information. Prominent architectures such as U-Net \cite{b14} and ResNet \cite{b15} have been widely employed in addressing the challenge of cloud removal. These architectures offer effective mechanisms for capturing spatial dependencies and contextual information. Most studies rely on both radar and optical data at one single timestamp or several timestamps \cite{b16}.

\section{METHOD}
\label{sec:method}

\subsection{Dataset}
\label{ssec:dataset}

In this study, we use publicly available Sentinel 1 and 2 data. LAI data is computed from Sentinel 2 data\footnote{\href{https://github.com/sentinel-hub/custom-scripts/tree/master/sentinel-2/lai}{https://github.com/sentinel-hub/custom-scripts/tree/master/sentinel-2/lai}} in different European countries. We pair this data with Sentinel 1 radar data captured no more than a day from the passage of Sentinel 2. Specifically, we used data from the VH and VV polarizations of Sentinel 1. Each sample corresponds to 3 consecutive Sentinel 2 LAI maps $S_2^T$, pixel-aligned with corresponding Sentinel 1 radar data images $S_1^T$, with $T \in  \{t-2,t-1,t\}$. Those three timestamps are separated by no more than a week. This data is processed into images of $256 \times 256$ pixels. We also have access to semantic masks ${S_{2}}_{mask}^T$ describing the nature of the element observed by Sentinel 2 (e.g. cloud, water, land, etc.). Masks are also computed from Sentinel 2 data using the Sen2Cor processor developed by ESA. Finally, the date of each observation is also available and plays a significant role in our method.

For testing purposes, we utilize $256 \times 256$ image sections randomly extracted from the previously mentioned locations. We create a first set from locations with consistently clear past observations and a second set from locations with consistently cloudy past observations. We validate our model on nearby squares during development. To further guard against validation overfitting, we incorporate an additional dataset from distinct locations, specifically from the Czech Republic and Italy. These datasets are subsequently referred to as "non-cloudy," "cloudy," and "unique areas," respectively.\footnote{we are grateful to World from Space for their help in acquiring, filtering and processing the data} The training dataset comprises 7,635 satellite images, including 250 clouded data, 254 non-clouded data, and 2,005 unique areas.

\begin{table*}
\caption{Ablation studies metrics}
\label{result_table}
\centering
\begin{tabular}{|l|c|c|c|c|c|c|c|}
\hline
\multicolumn{1}{|c|}{Input} & \multicolumn{1}{|c|}{Architecture} & \multicolumn{2}{c|}{Non cloudy} & \multicolumn{2}{c|}{Cloudy} & \multicolumn{2}{c|}{Unique areas} \\ 
\cline{3-8}
\multicolumn{1}{|c|}{data} &  & RMSE & R² & RMSE & R² & RMSE & R² \\
\hline
$S_1$                          & Encoder 1 & 0.163 & 0.434 & 0.279 & 0.152 & 0.260 & 0.125  \\
$S_1 + {S_2}_{masks} $         & Encoder 1 & 0.117 & 0.709 & 0.245 & 0.345 & 0.208 & 0.440 \\
$S_1 + {S_2}_{masks} + seas. $ & Encoder 1 & 0.111 & 0.735 & \textbf{0.233} & \textbf{0.409} & 0.198 & 0.491 \\
$S_2 + {S_2}_{masks} $         & Encoder 2 & 0.090 & 0.827 & 0.253 & 0.301 & 0.103 & 0.863 \\
$S_2 + {S_2}_{masks} + seas. $ & Encoder 2 & 0.067 & 0.903 & 0.344 &-0.294 & 0.111 & 0.839 \\
$All$                          & Final     & \textbf{0.058} & \textbf{0.930} & 0.238 & 0.383 & \textbf{0.101} & \textbf{0.867} \\
\hline
\end{tabular}
\end{table*}

\vspace{-0.2cm}
\subsection{Architecture}
\label{ssec:architecture}

\subsubsection{Description}
\label{ssec:description}

Our proposed architecture consists of three distinct components: two parallel encoders and one decoder (Figure \ref{archi_figure}).

The first encoder processes the input Sentinel 1 radar data, mask data, and seasonality information. The masks, which are pixel-wise one-hot-encoded maps, are fed to a point-wise convolution to reduce their dimensionality. The seasonality information, represented by sine and cosine values encoding the number of days elapsed since the beginning of the year, is fed through a multi-perceptron module before being broadcast into a feature map. This feature map is then concatenated with the other inputs and collectively passed through the U-net encoder. The second encoder is responsible for processing the input of LAI values at timestamps $t-2$ and $t-1$, as well as mask data and seasonality information. Similar to the first encoder, it follows an identical architecture, leveraging the same set of operations and modules.

The decoder component of our architecture receives the concatenated output of the two encoders as input. It is designed as a straightforward U-net structure, which facilitates the integration of the analyzed features from both the Sentinel 1 data and the Sentinel 2 data. By merging the analyzed features from both sources, our model leverages the complementary information provided by Sentinel 1 and 2 data.

\vspace{-0.2cm}
\subsubsection{Architecture design}
\label{ssec:design}

The U-net modules employed in our architecture have been widely recognized for their effectiveness in pixel-wise predictions, such as semantic segmentation tasks \cite{b14}. These modules offer several advantages, including a relatively small number of parameters, the incorporation of skip connections for enhanced convergence, and the ability to capture both local and global information. 

The overall design of our global architecture works as an inductive bias, directing the flow of information through the network. In line with this design, our approach prioritizes the shrinkage of multi-temporal data in the initial stages. This decision is motivated by the understanding that retaining multi-temporal data can lead to an unnecessary increase in the number of feature maps and potential redundancy when cloud cover is absent.

\vspace{-0.2cm}
\subsubsection{Intermediate supervision and pre-trained weights}

Given the complexity of our architecture, we adopted a two-step training approach. First, we trained each encoder separately by introducing a pixel-wise convolution layer and optimizing the mean squared error (MSE) loss for each encoder, with the ground truth label being the LAI. This initial training allowed us to obtain pre-trained weights for each encoder.

Subsequently, we incorporated these pre-trained weights into the entire architecture, utilizing intermediate supervision to optimize the overall model. Intermediate supervision involves optimizing the loss at intermediate stages, ensuring that the model maximizes the utilization of each input modality. This approach not only enables the model to make the most of the pre-trained weights' feature representation but also helps preserve and enhance the learned features during the early stages of training. The loss between the ground truth $LAI_t^{gt}$ and $pred = \{ LAI_t^{dec}, LAI_t^{enc_1}, LAI_t^{enc_2} \} $ is expressed as:

\vspace{-0.7cm}
\begin{align}
loss(LAI_t^{gt}, pred) &= MSE(LAI_t^{gt}, LAI_t^{dec}) \notag \\ 
&+ \alpha . MSE(LAI_t^{gt}, LAI_t^{enc_1}) \notag \\ 
&+ \beta . MSE(LAI_t^{gt}, LAI_t^{enc_2}) \notag
\end{align}

\vspace{-0.2cm}
$\alpha$ and $\beta$ are hyper-parameters to weigh down the importance of intermediate stages compared to the last stage. Note that in practice, the loss only takes into account pixels that are not covered by clouds, which we extract from ${S^t_2}_{mask}$.

\section{EXPERIMENTS}
\label{sec:typestyle}

\subsection{Implementation details}
\label{ssec:metrics}

The implementation was made in Pytorch. We trained with Adam for 100 epochs with an initial learning rate of 0.001. To enhance training stability and performance, we incorporated learning rate decay, applying two equally spaced decays with a decay factor of 0.2. For the loss parameters in the intermediate supervision, we used $\alpha = 0.1$ and $\beta=0.15$. Setting $\alpha  < \beta$ seemed to work better as previous LAI values provided more relevant information. The batch size was set to 32 and the training was carried out on a NVIDIA GeForce RTX 3090 GPU.

\vspace{-0.2cm}
\subsection{Results}
\label{ssec:results}

To directly compare whole-image input with pixel-wise prediction, we trained Multi Linear Regression (MLR) and Random Forest (RF) models using the same features but with only single-pixel data as a sample. The performance was notably poor: MLR achieved $R^2 = -0.0445$, RF had $R^2 = -0.0166$ on non-clouded data, while our method achieved $R^2 = 0.930$. This discrepancy with other LAI prediction research can be attributed to differences in evaluation settings; previous studies often used narrow, idealized regions, whereas our dataset represents real-world conditions.

In order to evaluate the effectiveness of our proposed method and gain deeper insights into the relationship between the input data and various components of the architecture, we conducted a series of ablation studies. These studies involved selectively training specific parts of the architecture and focusing on specific subsets of the input data. When training only on $S_1$ (resp. $S_2$), we drop the second (resp. first) encoder and the decoder, and we take the output of the intermediate stage for evaluation. See Table \ref{result_table}.

The high correlation observed between non-cloudy and unique area metrics provides reassurance regarding the performance of our model, both during training and at inference time on unseen locations. However, it is worth noting that the error rates were found to be twice as high in unique areas, indicating that the models excel primarily in trained areas. This suggests that the models might have learned biases specific to the training locations. Therefore, incorporating additional meta-information about the location, such as terrain variables or meteorological data, could potentially enhance the model's performance further. The positive impact of seasonality on the metrics further supports this idea.

The evaluation of our models on cloudy data revealed more mixed results. Specifically, we observed that $S_1$ provided more relevant information compared to $S_2$. Surprisingly, even our final model incorporating $S_1$ data exhibited poorer performance on cloudy data. We attribute this discrepancy to the dataset's inherent imbalance concerning cloudy data. The scarcity of cloudy data in the training set may have hindered the models' ability to optimize their reliance on $S_1$ data when faced with cloudy conditions. To address this challenge, an interesting approach would involve assigning a higher weight to the loss function for cloudy inputs during training.

\section{CONCLUSION}
\label{sec:conclusion}

Our study has demonstrated the effectiveness of deep learning techniques in predicting LAI by integrating global and local information from both radar and multispectral data sources at several timestamps. Moving forward, there is potential for further improvement by specifically training the model on cloudy data, emphasizing the importance of handling data with varying cloud cover. 

We also express our gratitude to World from Space for providing the data for this project and for their support.



\bibliographystyle{plainnat}
\bibliography{myBibFile}

\end{document}